\documentclass[10pt,twocolumn,letterpaper]{article}

\usepackage{egpaper}
\usepackage{times}
\usepackage{epsfig}
\usepackage{graphicx}
\usepackage{amsmath}
\usepackage{amssymb}

\usepackage[export]{adjustbox}
\usepackage{amsmath,amssymb}
\DeclareMathOperator*{\argmax}{argmax}
\newcommand{\matr}[1]{\mathbf{#1}}
\DeclareMathOperator{\IoU}{\matr{IoU}}
\usepackage{mathtools}
\usepackage{float}
\usepackage[ruled, linesnumbered]{algorithm2e}
\usepackage{multirow}
\DeclarePairedDelimiterX\set[1]\lbrace\rbrace{#1}
\newcommand{\squeezeup}{\vspace{-3mm}}

\newcommand{\expandup}{\vspace{+3mm}}
\usepackage{algorithmic}

\usepackage{hhline}
\usepackage{tabularx}
\usepackage{subfigure}
\makeatletter
\newcommand{\thickhline}{%
    \noalign {\ifnum 0=`}\fi \hrule height 1.2pt
    \futurelet \reserved@a \@xhline
}
\DeclareMathOperator{\Tr}{Tr}

\usepackage[pagebackref=true,breaklinks=true,letterpaper=true,colorlinks,bookmarks=false]{hyperref}

\iccvfinalcopy 

\def\iccvPaperID{4950} 

\ificcvfinal\fi
\begin{document}

\title{Learning Instance-Aware Object Detection Using Determinantal Point Processes}

\author{Nuri Kim\\
Seoul National University\\
{\tt\small nuri.kim@rllab.snu.ac.kr}
\and
Donghoon Lee\\
Seoul National University\\
{\tt\small donghoon.lee@rllab.snu.ac.kr}
\and
Songhwai Oh\\
Seoul National University\\
{\tt\small songhwai@snu.ac.kr}
}

\maketitle

\begin{abstract}
Recent object detectors find instances while categorizing candidate
regions.
As each region is evaluated independently, the number of candidate
regions from a detector is usually larger than the number of objects.  
Since the final goal of detection is to assign a single detection to
each object, a heuristic algorithm, such as non-maximum suppression
(NMS), is used to select a single bounding box for an object.
While simple heuristic algorithms are effective for
stand-alone objects, they can fail to detect overlapped objects. 
In this paper, we address this issue by training a network to distinguish
different objects using the relationship between
candidate boxes.
We propose an instance-aware detection network (IDNet), which can
learn to extract features from candidate regions and measure their
similarities.
Based on pairwise similarities and detection qualities, the
IDNet selects a subset of candidate bounding boxes using
instance-aware determinantal point process inference (IDPP).
Extensive experiments demonstrate that the proposed algorithm achieves
significant improvements for detecting overlapped objects compared to existing state-of-the-art detection methods on the PASCAL VOC and MS COCO datasets.\footnote{This paper is under consideration at Computer Vision and Image Understanding.}
\end{abstract}

\section{Introduction}
\begin{figure}[!t]{\centering\includegraphics[width=\linewidth]{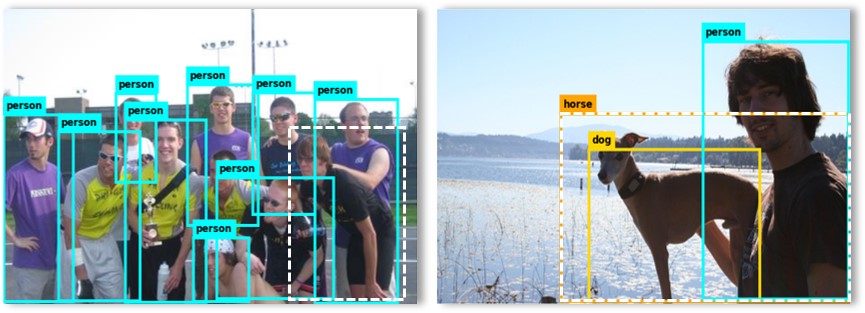}}
\caption{\textbf{Detection errors.} The errors are indicated by the dashed boxes. All examples are from baseline Faster R-CNN detector on PASCAL VOC.
\textbf{Left:} A crowded scene with people, where a \textit{person} is not detected.
\textbf{Right:} An image with duplicated detections from a single object, where a \textit{dog} is mistaken as a \textit{horse}. 
}
\label{fig:front}
\end{figure}

Object detection is one of the fundamental problems in computer
vision. 
Its goal is to detect objects by classifying and regressing bounding boxes in an image \cite{girshick2014rich, girshick2015fast,
  ren2015faster, redmon2016you, redmon2016yolo9000, liu2016ssd}.
It has received much attention because of its wide range of
applications, such as object tracking \cite{andriluka2008people},
surveillance \cite{tian2005robust}, and face detection
\cite{ranjan2017hyperface}.  
Most of the state-of-the-art detectors show significant performance
improvements based on deep convolutional neural networks.
Despite the advances in object detection, it is still difficult to
assign correct detections for all objects in an image since
detectors do not distinguish different object instances in the same
class as it only focuses on an instance-agnostic task, i.e., object
category classification.
This issue becomes critical when objects are overlapped.
As shown in the left image of Figure~\ref{fig:front}, the person
on the right is not detected due to the overlapped bounding boxes in proximity.

In order to address this issue, we propose an instance-aware detection network (IDNet),
which learns to differentiate representations of different objects.
IDNet learns the similarity among
bounding boxes during training and selects a subset of boxes based on the learned similarity during inference.
Specifically, IDNet learns to compare
appearances of bounding boxes while considering their spatial arrangements.

IDNet uses an existing detector, such as Faster R-CNN, as a component
to obtain candidate bounding boxes. 
Given candidate boxes, IDNet extracts features of all candidates
using a CNN branch, named a region identification network (RIN), which
aims to increase the probability of selecting an optimal subset of detections.
To this end, IDNet is trained not only with the softmax loss and smooth L1 loss \cite{ren2015faster}, but also with novel losses based on determinantal
point processes (DPPs) \cite{kulesza2012determinantal}.
A DPP is used in various machine learning fields, such as document and video summarization \cite{chao2015large, zhang2016video, lin2012learning}, sensor placement \cite{krause2008near}, recommendation systems \cite{zhou2010solving} and multi-label classification \cite{xie2017deep}, to select a desirable subset from a set of candidates. 
Using the property of repulsiveness in DPPs,
we design an instance-aware detection loss (ID loss), which learns to
increase the probability of selecting an instance-aware subset from detection candidates.

Another source of the detection error is multiple detections of different classes for a single object.
This error has been known to be one of the persistent problems for
instance-agnostic detectors, such as Faster R-CNN \cite{ren2015faster}. 
For example, as shown in the right image of Figure~\ref{fig:front}, 
there are two
bounding boxes categorized as a \textit{dog} and a \textit{horse} for the same object. 
Since the objective of a detector is to find a single bounding box for
a single object instance, we propose the sparse-score loss (SS loss) using DPPs to make
IDNet assign a single bounding box for a single object, considering
all categories.
In particular, we formulate the SS loss to remove duplicated
bounding boxes by training IDNet to have low confidence
scores for bounding boxes with incorrect class labels.
After training, our algorithm efficiently finds a subset of candidate
detections using the log-submodular property of DPPs
\cite{kulesza2012determinantal}.

Experimental results show that IDNet is more robust for
detecting overlapped objects against the
baseline detectors, such as Faster R-CNN \cite{ren2015faster}
and learning detection with diverse proposals (LDDP) \cite{azadi2017learning}, on PASCAL VOC
\cite{everingham2010pascal} and MS COCO \cite{lin2014microsoft}.
Our IDNet achieves 5.8\% mAP improvement on PASCAL VOC 2007 and 2.5\% mAP improvement on PASCAL VOC 0712 over Faster R-CNN when tested on the VOC crowd set, which consists of images with overlapped objects.
For COCO, the performance is improved by 1.3\% AP when tested on the COCO crowd set.

The main contributions of this paper are summarized as follows:
(1) Two novel losses, the sparse-score loss and
the instance-aware diversity loss, are proposed for instance-aware detection;
(2) To the best of our knowledge, this work is the first approach that
trains a neural network to learn quality and diversity terms of a DPP for object detection;
(3) The proposed algorithm outperforms baseline detectors for detecting overlapped objects.

\section{Related Work}
\paragraph{Class-aware detection algorithms.}
The goal of class-aware or multi-class object detection is to
localize objects in an image while predicting the category of each
object. 
These systems are usually composed of region proposal networks and
region classification networks \cite{girshick2015fast,ren2015faster,liu2016ssd}. 
To improve detection accuracy, a number of different optimization
formulations and network architectures have been proposed \cite{ren2015faster, kong2016hypernet, azadi2017learning, redmon2016you, liu2016ssd, redmon2016yolo9000, dai2016r}.
Ren \etal \cite{ren2015faster} use convolutional networks,
called region proposal networks, to get region proposals and combine
it with Fast R-CNN. 
Kong \etal \cite{kong2016hypernet} utilizes each layer's feature for detecting small objects in an image. 
A real-time multi-class object detector is proposed by combining
region proposal networks and classification networks in
\cite{redmon2016you}. 
Liu \etal \cite{liu2016ssd} improve the performance of
\cite{redmon2016you} using multiple detectors for each
convolutional layer.  
To increase network efficiency, fully connected layers are replaced by
convolution layers in \cite{dai2016r}.
Redmon \etal \cite{redmon2016yolo9000} extend \cite{redmon2016you} by
classifying thousands of categories using the hierarchical structure
of categories in the dataset.  

DPPs have been used to improve detection qualities before.
Azadi \etal \cite{azadi2017learning} propose to suppress background
bounding boxes, while trying to select correct detections.  
However, this method focuses on ing detection scores and uses a
fixed visual similarity matrix based on WordNet \cite{miller1995wordnet}, while our algorithm
learns the similarity matrix from data.

\paragraph{Instance-aware algorithms.}
Instance-aware algorithms have been developed to provide finer solutions
in different problem domains.
Instance-aware segmentation aims to label instances at the pixel level
\cite{dai2016instance, ren2017end}. 
Li \etal \cite{dai2016instance} propose a cascade network which finds
each instance stage by stage. 
Similar to RIN, a network in \cite{dai2016instance} finds features of
each instance. 
Ren \etal \cite{ren2017end} use a recurrent neural network to
sequentially find each instance. 
A face detector which takes keypoints of faces as an input is
suggested in \cite{li2016face}. 
The dataset for this application contains face labels for
identifying different faces, while the standard object detection datasets
only have a small number of categories.

In object detection, Wang \etal \cite{wang2017repulsion} introduce a repulsion loss to improve localization of instances. However, their approach is limited to a single-class detection problem and uses NMS~\cite{felzenszwalb2010object} as a post-processing method. 
Lee \etal \cite{lee2016individualness} provide an
inference method to find an optimal subset of detection candidates for pedestrian detection
considering the individualness of each detection candidate. 
However, this approach tackles a single-class detection problem and
uses features computed from a network pre-trained on the ImageNet
dataset \cite{deng2009imagenet}, instead of training the network for
the desired purpose. 
Our method tackles a challenging multi-class detection task
by learning distinctive features of object instances from data. 

Recently, a detector which learns the structural relationship between
objects is proposed in \cite{liu2018structure}, where the detection
score of an object is scaled by considering scene context and
relationship between objects. 
Liu \etal \cite{liu2018structure} show that training with a
structural relationship can implicitly reduce redundant detection
boxes, while our method explicitly suppresses the scores of duplicated
detection boxes. 
Hu \etal \cite{hu2017relation} utilize a modified attention module
\cite{vaswani2017attention} for learning a relationship between
bounding boxes. 
The module scales the scores using the instance relationship similar
to ours. 
However, this method uses the standard softmax loss and smooth L1
loss, while our IDNet tackles this problem by training a detector with
novel losses. 

\section{Proposed Method}

\begin{figure*}[!t]
\begin{center}
    \makebox[\textwidth]{\includegraphics[width=0.9\textwidth]{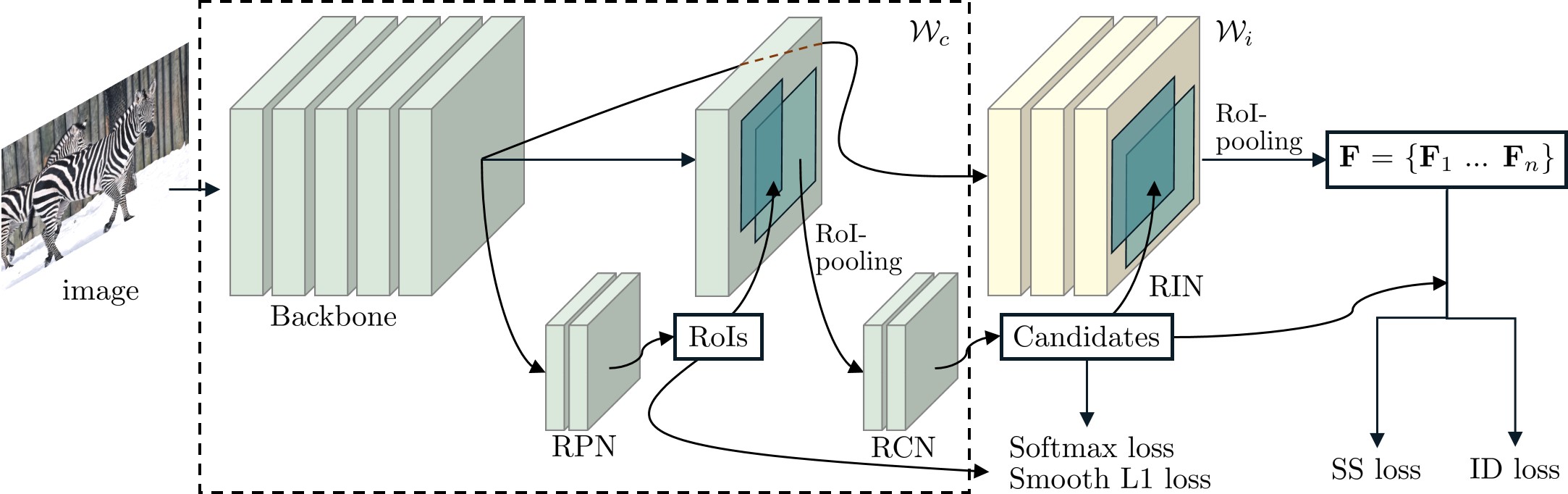}} 
    \caption{
\textbf{Pipeline of the instance-aware detection network (IDNet).} 
The dashed box indicates the weights of a backbone, RPN, and RCN
($\mathcal{W}_c$). 
The other weights are named as $\mathcal{W}_i$. 
Using the features extracted from RIN and the detection quality, 
a probability of each bounding box to be
selected can be calculated. 
IDNet is trained with the proposed SS loss and ID loss, as well as the
softmax and smooth L1 losses from Faster R-CNN \cite{ren2015faster}. 
The SS loss is used to suppress duplicated candidate boxes, and the ID
loss is used to learn the similarity between candidate boxes. 
The smooth L1 loss is for regressing bounding boxes to exact locations
of objects while classifying the objects in the boxes using the
softmax loss.
}  

\label{fig:net} 
\end{center}
\end{figure*}

An overview of the proposed IDNet is shown in Figure~\ref{fig:net}. 
IDNet is composed of a region proposal network (RPN), a
region classification network (RCN) and a region identification
network (RIN). 
Based on image feature maps from the backbone network, RPN predicts
region proposals, i.e., the region of interests (RoIs). 
Then, a RoI pooling layer pools regional features from feature maps for each RoI.
Using the regional features, RCN classifies the regions into multiple categories while localizing the regions. 
RIN computes instance features of candidates, which are used by DPPs.\footnote{RIN consists of seven convolutional layers and three fully connected layers. The detailed structure of RIN is described in the appendix.}

\subsection{Determinantal Point Processes for Detection}

Suppose that there are $n$ candidate bounding boxes, 
$\mathcal{Y} = \set{\textbf{b}_1 , \textbf{b}_2,  ... , \textbf{b}_n}$,
where $\textbf{b}_i$ is the $i$th bounding box.
A determinantal point process (DPP) defines a probability distribution over
subsets of $\mathcal{Y}$ as follows \cite{kulesza2012determinantal}.
If $\textit{\textbf{Y}}$ is a DPP, then
\begin{align}\small
\begin{aligned}[t] \label{eq:likeli}
 \mathcal{P}_\matr{L}(\textit{\textbf{Y}}=Y) 
= \frac{\det (\matr{L}_Y)}{\sum_{Y^{'} \subseteq \mathcal{Y}} \det(\matr{L}_{Y^{'}})}
= \frac{\det (\matr{L}_Y)}{\det (\matr{L}+\matr{I})},
\end{aligned}
\end{align}
where $Y \subseteq \mathcal{Y}$, 
a kernel matrix $\matr{L} \in \mathbb{R}^{n \times n}_{+}$ is a real
symmetric positive semi-definite matrix, an indexed kernel matrix
$\matr{L}_Y \in \mathbb{R}^{|Y| \times |Y|}_{+}$ is a submatrix of
$\matr{L}$ indexed by the elements of $Y$, and 
$\matr{I} \in \mathbb{R}^{n \times n}$ is an identity matrix. 
The kernel matrix $\matr{L}$ can be decomposed as $\matr{V}\matr{V}^T$,
where $\matr{V} \in \mathbb{R}^{n \times r}$ is a feature matrix for
$n$ candidate bounding boxes. Each row of $\matr{V}$ is extracted from RIN and normalized to construct the matrix.
Similar to the kernel matrix, the indexed kernel matrix  $\matr{L}_Y$
can be decomposed as $\matr{V}_Y\matr{V}_Y^T$. 

Let $\textbf{q}_i$ be the detection score for the $i$th bounding box.
Then, $\textbf{q} = [\textbf{q}_1, ..., \textbf{q}_n]$ is the detection
quality for all $n$ detection candidates.
The feature $\textbf{F}_i$ for $\textbf{b}_i$ is extracted from the RIN.
Let $\matr{V}_i = \textbf{F}_i/\|\textbf{F}_i\|_2$ be a normalized
feature, where $i \in \{1,...,n\}$.
Using candidate bounding boxes, the intersection over union between
$\textbf{b}_i$ and $\textbf{b}_j$ can be calculated by
${\IoU}_{ij} = {\#(\textbf{b}_i \cap \textbf{b}_j)} 
  /{\#(\textbf{b}_i \cup \textbf{b}_j)}$, where $\#\text{(A)}$ is the number of pixels in A,
and we construct a matrix $\IoU$ by setting $[\IoU]_{ij} = {\IoU}_{ij}$.
A similarity matrix $\matr{S}$ is constructed as
$\matr{S} = \lambda \cdot \matr{V}\matr{V}^T + (1 - \lambda) \cdot \IoU$, 
where $\lambda \in [0,1]$.  
Using  the similarity
matrix $\matr{S}$ and the detection quality $\textbf{q}$, the kernel matrix for a DPP \cite{kulesza2012determinantal} can be formed as 
$\matr{L} = \matr{S} \odot \textbf{q}\textbf{q}^T$, where $\odot$ is the element-wise multiplication.\footnote{
Notations in this paper are summarized in
the appendix.
}
 
If the similarity $\matr{S}$ and detection quality $\textbf{q}$ are
correctly assigned, a subset which maximizes (\ref{eq:likeli}) is a
collection of the most distinctive detections due to the property of
the determinant in a DPP \cite{kulesza2012determinantal}.
Since IDNet is trained to maximize the probability (\ref{eq:likeli})
of the ground-truth detections, IDNet learns the most distinctive
features and correctly scaled detection scores to separate difference
object instances in order to compute $\matr{S}$ and
$\textbf{q}$.  

\subsection{Learning Detection Quality} \label{learn_quality}

As RCN classifies each RoI independently, multiple detections 
with different categories often have high detection scores.
For example, a detector would report a \textit{horse} nearby a \textit{dog} as they are visually similar.
Then, conventional post-processing methods, such as NMS, are typically suppress bounding boxes in each class.
While heuristic post-processing algorithms are effective for removing duplicated bounding boxes in each category, 
these algorithms cannot remove duplicated boxes with different categories.
In this case, even if there is a true bounding box for the \textit{dog}, the \textit{horse} bounding box cannot be removed.
To alleviate this issue, we propose the sparse-score loss (SS loss) to
detect an object with the correct class label by removing the other
candidate boxes with incorrect categories. 

We first select categories with top $m$ detection scores among $n_c$
categories from each RoI. 
We assume that the selected categories are composed of visually
similar categories from the correct category. 
By suppressing the scores of visually similar categories except for
the bounding boxes of a top-1 category, we can obtain a single
bounding box with a correct category for an object. 
Let $\mathcal{Y}_m$ be all bounding boxes of  top-$m$ categories from
all RoIs and $Y_{pos}$ be a set of positive boxes, i.e., bounding
boxes with a top-1 category in each RoI. 
Then, we define the SS loss as the negative log-likelihood of
(\ref{eq:likeli}) as follows: 
\begin{align}\small
\begin{aligned}[t] \label{eq:ss_loss}
    &\mathcal{L}_{SS}(Y_{pos}, \mathcal{Y}_{m}) \\
    &=  -\log\left(\sum_{Y
      \subseteq
      Y_{pos}}{\mathcal{P}_{\matr{L}_{\mathcal{Y}_m}}(Y)}\right) \\
    &= -\log\left(\sum_{Y \subseteq Y_{pos}} 
    \frac{\det(\matr{L}_Y)}{\det(\matr{L}_{\mathcal{Y}_{m}}+\matr{I}_{\mathcal{Y}_{m}})}\right)
       \\
&= - \log\det (\matr{L}_{Y_{pos}}+\matr{I}_{Y_{pos}}) + \log \det
      (\matr{L}_{\mathcal{Y}_{m}}+\matr{I}_{\mathcal{Y}_{m}}), 
\end{aligned}
\end{align}
where $\sum_{Y \subseteq Y_{pos}}{\det(\matr{L}_Y)}= \det(\matr{L}_{Y_{pos}}+\matr{I}_{Y_{pos}})$.
This loss function increases detection scores of bounding boxes in the
positive set, $Y_{pos}$. 
In other words, this loss suppresses scores of all subsets which have
at least one non-positive bounding box. 
We would like to note that the normalization term for a DPP is
included for numerical stability during training.

We also use two softmax losses for classification ($\mathcal{L}_b$ for binary classification and $\mathcal{L}_m$ for multi-class classification), and two smooth L1 losses ($\mathcal{L}_r$) for the RoI regression and candidate box regression \cite{ren2015faster}.
Note that the losses are the same as \cite{ren2015faster} since we adopt Faster R-CNN as a baseline. We call the summation of all above losses as a multi-task loss.

Suppose RPN predicts the objectness probability $p_i$ and location shifts $l_i$, where $i$ is the index of RoIs in a mini-batch. RCN predicts $c_j$ of categories and location shifts $t_j$, where $j$ is the index of candidate boxes. The target location shift for the $i$th RoI and $j$th candidate box are $l_i^{*}$ and $t_j^{*}$, respectively. Additionally, $p_i^*$ and $c_j^*$ are the ground truth category label for a RoI and an candidate box, respectively.
Then, the multi-task loss is expressed as follows:  
\begin{align}\small
\begin{aligned}[t] \label{eq:cls_reg}
&\mathcal{L}_{MT}(\{p_i\},\{l_i\},\{c_j\},\{t_j\}) \\ 
&= \sum_{i}\mathcal{L}_{b}(p_i, p_i^*) + \sum_{i}\textbf{1}_{p_i^*>0} \cdot \mathcal{L}_{r}(l_i, l_i^*) \\
& + \sum_{j}\mathcal{L}_{m}(c_j, c_j^*) +  \sum_{j}\textbf{1}_{c_j^*>0} \cdot \mathcal{L}_{r}(t_j, t_j^*),
\end{aligned}
\end{align}
where $\textbf{1}_{c_j^*>0}$ is an indicator function, which outputs 1 when the $j$th candidate box has a non-background label.

With all losses defined as above, the weights for a backbone, RPN, and RCN, which are denoted by
$\mathcal{W}_c$ in Figure~\ref{fig:net}, can be learned by optimizing:
\begin{align}\small
\begin{aligned}[t]
\min_{\mathcal{W}_{c}} \lambda_{ss} \cdot \mathcal{L}_{SS}(Y_{pos}, \mathcal{Y}_{m}) + \mathcal{L}_\text{MT}(\{p_i\},\{l_i\},\{c_i\},\{t_i\}),
\end{aligned}
\end{align}
where $\lambda_{ss}$ is used to balance the SS loss with the multi-task loss. The similarity matrix $\matr{S}$ is fixed while calculating the gradient of the SS loss, since $\mathcal{W}_{i}$ is freezed while optimizaing $\mathcal{W}_{c}$. 

\subsection{Learning Instance Differences}

An instance-agnostic detector solely based on object category
information often fails to detect objects in proximity. 
For accurate detections from real-world images with frequent
overlapping objects, it is crucial to distinguish different object
instances.
To address this problem, we propose the instance-aware detection loss
(ID loss). 
The objective of this loss function is to obtain similar features from
the same instance and different features from different instances. 
This is done by maximizing the probability of a subset of the most
distinctive bounding boxes. 

Let $\mathcal{Y}_s$ be a set of all candidate bounding boxes which
intersect with the ground truth bounding boxes. 
Let $Y_{rep} \subseteq \mathcal{Y}_{s}$ be a set of the most
representative boxes, i.e., candidate boxes which are closest to the
ground truth boxes obtained by the Hungarian algorithm~\cite{kuhn1955hungarian}.
Then, the ID loss for all objects is defined as follows:
\begin{align}\small
\begin{aligned}[t]
    \mathcal{L}_{ID}^{all}(Y_{rep}, \mathcal{Y}_{s}) &= -\log(\mathcal{P}_{\matr{L}_{\mathcal{Y}_s}}(Y_{rep})) \\ &= - \log \det (\matr{L}_{Y_{rep}}) + \log \det (\matr{L}_{\mathcal{Y}_{s}}+\matr{I}_{\mathcal{Y}_{s}}).
\end{aligned} \label{eq:id_all}
\end{align}
Due to the determinant, it increases the cosine distance between $V_i$
and $V_j$ if $i$ and $j$ are from different instances. 
As we select boxes nearby the ground truth bounding boxes to construct
$\mathcal{Y}_{s}$, the network can learn what bounding boxes are
similar or different. 

In addition to (\ref{eq:id_all}), we set an objective which
focuses on differentiating instances from the same category. 
For category $C_k$, $\mathcal{Y}_{C_k}$ is candidate boxes in the $k$th category and 
${Y}_{C_k} \subseteq \mathcal{Y}_{C_k}$ is a set of candidate boxes
which are closest to the ground truth boxes.
${Y}_{C_k}$ is also obtained by the Hungarian algorithm~\cite{kuhn1955hungarian}.
The category-specific ID loss is defined as follows:
\begin{align}\small
\begin{aligned}[t]
    &\mathcal{L}_{ID}^{ic}({Y}_{C_k}, \mathcal{Y}_{C_k}) \\&=  -\log(\mathcal{P}_{\matr{L}_{\mathcal{Y}_{C_k}}}(Y_{C_k})) \\ &=- \log \det (\matr{L}_{{Y}_{C_k}}) + \log \det (\matr{L}_{\mathcal{Y}_{C_k}}+\matr{I}_{\mathcal{Y}_{C_k}}).
\end{aligned}
\end{align}
It provides an additional guidance signal to train the network since it
is more difficult to distinguish similar instances from the same
category than instances from different categories. We find an improvement when we use both $\mathcal{L}^{all}_{ID}$ and $\mathcal{L}^{ic}_{ID}$, compared to cases when only one of them is used.
Finally, the ID loss is defined as:
\begin{align}\small
\begin{aligned}[t]
    &\mathcal{L}_{ID}(Y_{rep}, \mathcal{Y}_{s}, {Y}_{C_k}, \mathcal{Y}_{C_k}) \\&= \mathcal{L}_{ID}^{all}(Y_{rep}, \mathcal{Y}_{s}) +\frac{1}{K} \cdot \sum_{k=1}^{K}  \mathcal{L}_{ID}^{ic}({Y}_{C_k}, \mathcal{Y}_{C_k}).
\end{aligned}
\end{align}
The goal of the ID loss is to find all instances while discriminating
different instances as shown in Figure~\ref{fig:front}. 
While the ID loss aims to distinguish instances, the multi-task loss tries to classify categories. The difference between their goals makes a network perform worse when both losses are used simultaneously. To alleviate the problem, we trained weights of RIN ($\mathcal{W}_i$ in Figure~\ref{fig:net}) separate from $\mathcal{W}_{c}$.
Given a set of candidate bounding boxes and subsets of them, weights
of RIN can be learned by optimizing:\footnote{The gradients of the SS, ID losses are derived in the appendix.} 
\begin{align}\small
\begin{aligned}[!h]
\min_{\mathcal{W}_{i}} \mathcal{L}_{ID}(Y_{rep}, \mathcal{Y}_{s}, {Y}_{C_k}, \mathcal{Y}_{C_k}).
\end{aligned}
\end{align}
Note that while calculating the gradient of the ID loss, the detection quality ($\textbf{q}$) is fixed, as $\mathcal{W}_{c}$ is freezed while optimizaing $\mathcal{W}_{i}$.

\subsection{Inference}  

Given a set $\mathcal{Y}$ of candidate bounding boxes, the similarity
matrix $\matr{S}$ and the detection quality $\textbf{q}$,
Algorithm~\ref{dpp_infer} (IDPP) finds the most representative
subset of bounding boxes. 
The problem of finding a subset that maximizes the probability is NP-hard \cite{kulesza2012determinantal}.
Fortunately, due to the log-submodular property of DPPs 
\cite{kulesza2012determinantal}, we can approximately solve the
problem using a greedy algorithm, such as
Algorithm~\ref{dpp_infer}, which iteratively adds an index
of a detection candidate until it cannot make the cost of a new
subset higher than that of the current subset \cite{azadi2017learning}, where the cost of a set 
$Y$ is $\log(\prod_{i \in Y} \textbf{q}_i^2 \cdot \det (\matr{S}_{{Y}}))$.
\begin{algorithm}[!t]
\caption{\small Instance-Aware DPP Inference (IDPP).}
\begin{algorithmic}[1]\itemindent=-0.9pc \small
\label{dpp_infer}
\STATE{$Y^* = \emptyset$}
  \WHILE{$\mathcal{Y} \neq \emptyset$}\itemindent=-0.9pc
    \STATE{$j^* = {\argmax}_{j \in \mathcal{Y}} \log(\prod_{ i \in Y^* \cup \{j\} } \textbf{q}_i^2 \cdot \det (\matr{S}_{{Y^* \cup \{j\}}})) $}
    \STATE{$Y = Y^* \cup \{j^*\} $}
    \IF{$\text{Cost}(Y) > \text{Cost}(Y^*)$}\itemindent=-0.9pc
        \STATE{$Y^* \gets Y $}
        \STATE{delete $j^*$ from $\mathcal{Y}$}
    \ELSE\itemindent=-0.9pc
        \RETURN{$Y^*$}
    \ENDIF
  \ENDWHILE
  \RETURN {$Y^*$}
\end{algorithmic} 
\end{algorithm}

\section{Experiments}
\paragraph{Datasets and baseline methods.}
We comprehensively evaluated IDNet on PASCAL VOC \cite{everingham2010pascal} and MS COCO 2014 \cite{lin2014microsoft}, which include 20 and 80 categories, respectively.

To demonstrate that our IDNet is effective for detecting overlapped objects, we have constructed the VOC crowd set from the VOC 2007 \texttt{test} set and the COCO crowd set from the COCO \texttt{val} set, respectively.
The crowd sets contain at least one overlapped object in an image. 
Unless otherwise specified, we define overlapped objects as those who overlap with another object over 0.3 IoU in all experiments.
We name the crowd set on VOC 2007 as $\text{VOC}_{crd}$, where the number of images is 283.
The COCO crowd set consists of 5,471 images, which is called $\text{COCO}_{crd}$. 
The indices of crowd sets will be made publicly available.

Since the goal of our algorithm is to discriminate instances with
given candidate bounding boxes, we adopt Faster R-CNN as a proposal
network to get candidate detections, but other proposal networks can be used in our framework. 
We implement baseline methods, Faster R-CNN \cite{ren2015faster} and LDDP \cite{azadi2017learning}, to compare with our algorithm.
Since there are few methods tested on the crowd sets, we choose the two baselines for fair comparison.
Note that our baseline implementation achieves a reasonable performance of 71.4\% mAP when trained with VOC 2007 using VGG-16 as a backbone, considering that the performance in the original paper~\cite{ren2015faster} is 69.9\% mAP.

We use different inference algorithms for each method. Unless otherwise stated, Faster R-CNN uses NMS, LDDP uses LDPP, and IDNet uses IDPP as an inference algorithm. LDPP is an inference algorithm proposed in LDDP \cite{azadi2017learning}, which uses a fixed class-wise similarity matrix while our IDPP uses the instance-aware features extracted from RIN.

\paragraph{Implementation details.}
All baseline methods and our IDNet are implemented based on the Faster R-CNN in Tensorflow \cite{abadi2016tensorflow}, where the most parameters, such as a learning rate, optimizer, data augmentation strategy, and batch size, are the same as the original paper \cite{ren2015faster}.
In our method, we use backbone networks, e.g., VGG-16 and ResNet-101, pre-trained on the ImageNet \cite{deng2009imagenet} and the RIN module is initialized with Xavier initialization~\cite{glorot2010understanding}. The RIN shares the parameters in a backbone, such as the layers until the $\texttt{conv2}$ of VGG-16 \cite{simonyan2014very} and the $\texttt{conv1}$ of ResNet-101 \cite{he2016deep}, to conserve memory.
We set $m$ to five for the VOC and ten for the COCO, since VOC has around five categories in the super-category and COCO has ten categories in the super-category on average.
We set the ratio between the spatial similarity and visual similarity ($\lambda$) to 0.6, which is a similar value compared with \cite{zhang2016video, lee2016individualness}.
Since the performance of a detector is poor during the early stage of training, top-$m$ bounding boxes do not contain similar categories. Thus, we set $\lambda_{ss}$ to zero during the early stage of training. $\lambda_{ss}$ is increased to 0.01 after the early stage.
The early stages are chosen around 60\% of total training iterations. We use 40k iterations for VOC 2007, 70k for VOC 0712, and 360k for COCO. 
Additionally, we set the size of $\matr{F}_i$ to 256 as it performs the best.
More implementation details can be found in the appendix.
\paragraph{Evaluation metrics.}
For evaluation, we use the mean average precision (mAP). 
We report mAP which considers detection candidates over IoU 0.5 as correct objects for VOC.
For COCO, we evaluate performance with three types of mAPs in the standard MS COCO~\cite{lin2014microsoft} protocols: $\text{AP}$,
$\text{AP}_{50}$, and $\text{AP}_{75}$. 
$\text{AP}$ reports the average values of mAP at ten different IoU thresholds from .5 to .95,
$\text{AP}_{50}$ reports mAP at IoU 0.5, and 
$\text{AP}_{75}$ reports mAP at IoU 0.75. 
A high score in $\text{AP}_{75}$ requires better localization of detection boxes.

\begin{figure}[!t]{
\centering\resizebox{\linewidth}{!}{\centering\includegraphics[width=\linewidth]{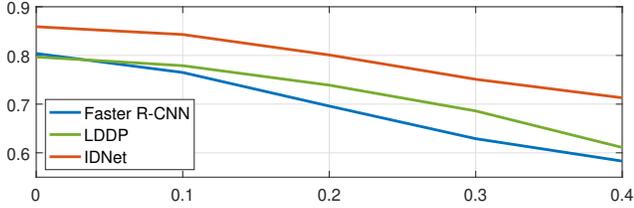}}
\caption{\textbf{Recall curves of Faster R-CNN, LDDP, and IDNet on VOC 2007.} The results are evaluated at different overlap IoU thresholds, from .0 to .4. Our proposed IDNet has a higher crowd recall and effectively detects object with high overlaps.}
\label{fig:result_overlap} }
\end{figure} 

\subsection{PASCAL VOC} \label{sec:pascal_voc}

For VOC 2007, we train a network with VOC 2007 \texttt{trainval}, which contains 5k images. For VOC 0712, we train a network with VOC 0712 \texttt{trainval} set, which includes 16k images. All methods are tested on VOC 2007 \texttt{test} set, which has 5k images.
After training IDNet with the SS loss and the multi-task loss, we train RIN to learn differences of instances with the ID loss for 30k iterations for VOC 2007, and 20k iterations for VOC 0712. While training RIN, the parameters in other modules except RIN are frozen.
A VGG-16 backbone is used for all tested methods for PASCAL VOC.

Since IDNet is effective for overlapped objects, we report recall which is calculated as a ratio of detected objects among the overlapped objects (Figure~\ref{fig:result_overlap}). 
For calculating recall, we check that there are detected objects among the objects overlapped with another object above a fixed IoU threshold.
After calculating the probability of detecting overlapped objects in each category, the results are
averaged over categories. 
The recall is a better performance measure than mAP for showing the robustness to overlap. This is because the recall is calculated only for overlapped objects, while the mAP is calculated for all objects in an image containing at least a single overlapped object. 

In Figure~\ref{fig:result_overlap}, recall for the objects with overlap over 0.4 is
increased from 0.58 (Faster R-CNN) to 0.71 (IDNet), which is an impressive improvement.
For all overlap regions, recall is higher than baseline methods and as the overlap ratio gets higher, the performance gap between Faster R-CNN and IDNet gets bigger.
The results show that IDNet is effective for detecting objects in proximity.

\begin{table}[!t]
\centering
\caption{\textbf{Detection results on VOC 2007 \texttt{test} set and VOC crowd set.} Legend: \textbf{07:} VOC 2007 \texttt{trainval} set, \textbf{07+12:} VOC 0712 \texttt{trainval} set. All methods are trained using a VGG-16 backbone network.}
\resizebox{\linewidth}{!}{
\begin{tabular}{ccc|cc} \thickhline
\multirow{2}{*}{Method} & \multirow{2}{*}{Inference} & \multirow{2}{*}{Train} &  \multicolumn{2}{c}{mAP} \\ \cline{4-5}
 & & & $\text{VOC}_{all}$ & $\text{VOC}_{crd}$\\ \thickhline
Fast R-CNN~\cite{girshick2015fast} & NMS & 07 & 66.9  & - \\
SSD300~\cite{liu2016ssd} & NMS & 07& 68.0  & - \\
Faster R-CNN~\cite{ren2015faster} & NMS & 07 & 71.4& 56.0 \\
LDDP~\cite{azadi2017learning} & LDPP & 07 & 70.9 & 57.7 \\  
$\text{IDNet}^*$ & IDPP & 07 & \textbf{71.9} &\textbf{61.8} \\ \hline
Fast R-CNN~\cite{girshick2015fast} & NMS & 07+12 & 70.0  & - \\
SSD300~\cite{liu2016ssd} & NMS & 07+12 & 74.3  & - \\
Faster R-CNN~\cite{ren2015faster} & NMS & 07+12 & 75.8 & 62.0 \\
LDDP~\cite{azadi2017learning} & LDPP & 07+12 & 76.4 &  63.1 \\ 
$\text{IDNet}^*$ & IDPP & 07+12 & \textbf{76.6} & \textbf{64.5} \\
\thickhline 
\end{tabular}}
\label{tab:pascal_all}  
\end{table}

\begin{table*}[!t]
\centering
\caption{\textbf{Detection results on COCO \texttt{val} set and COCO crowd set.} All methods are trained with COCO \texttt{train} set.}
\resizebox{\textwidth}{!}{
\begin{tabular}{ccc|cc|cc|cc} \thickhline
\multirow{2}{*}{Method} & \multirow{2}{*}{Inference} &
 \multirow{2}{*}{Backbone} &
 \multicolumn{2}{c|}{$\text{AP}$} &  \multicolumn{2}{c|}{$\text{AP}_{50}$} &  \multicolumn{2}{c}{$\text{AP}_{75}$} \\ \cline{4-9}
 & & & $\text{COCO}_{all}$ & $\text{COCO}_{crd}$ & $\text{COCO}_{all}$ & $\text{COCO}_{crd}$ & $\text{COCO}_{all}$ & $\text{COCO}_{crd}$\\ \thickhline
Faster R-CNN~\cite{ren2015faster} & NMS & VGG-16 & 26.2 & 19.2& 46.6 & 36.9 & 26.9 & 18.4 \\
LDDP~\cite{azadi2017learning} & LDPP & VGG-16& 26.4 & 19.6 &46.7& 37.9& 26.8 &18.6\\  
IDNet & IDPP & VGG-16 & \textbf{27.3}& \textbf{20.5}& \textbf{47.6}& \textbf{38.2} & \textbf{28.2} & \textbf{20.0} \\
\hline
Faster R-CNN~\cite{ren2015faster} & NMS &  ResNet-101 & 31.5& 23.5 & 52.0 & 42.5 &33.5&23.0\\
LDDP~\cite{azadi2017learning} & LDPP &  ResNet-101& 31.4 &23.8 & 51.7& 43.0 & 33.4 & 23.4 \\ 
IDNet & IDPP & ResNet-101& \textbf{32.7} & \textbf{24.4} & \textbf{53.1} & \textbf{43.4}& \textbf{34.8} & \textbf{24.4}\\
\thickhline 
\end{tabular} 
} 
\label{tab:coco_all}  
\end{table*}
To demonstrate that our IDNet is effective for detecting overlapped
objects on the standard mAP, we tested Faster R-CNN, LDDP and our $\text{IDNet}^*$\footnote{$\text{IDNet}^*$ is a version of IDNet only using ID loss.} on the VOC crowd
set ($\text{VOC}_{crd}$ in Table~\ref{tab:pascal_all}).
The IDNet  shows impressive improvements compared to Faster R-CNN with an improvement of 5.8\% mAP for VOC 2007 and 2.5\% for VOC 0712.
We also observe improvements over LDDP: 4.1\% improvement in mAP for VOC 2007
and 1.4\% improvement for VOC 0712.
Next, when we evaluated mAP for $\text{VOC}_{all}$, the mAP compared with baseline methods
is increased for both VOC 2007 and VOC 0712
(Table~\ref{tab:pascal_all}). 

\subsection{MS COCO}
MS COCO is composed of 80k images in the \texttt{train} set and
40k images in the \texttt{val} set.
After training a network with the SS loss and the multi-task loss, we
train the RIN module with the ID loss for 20k additional iterations. 

In Table~\ref{tab:coco_all}, we report the results using multiple APs for COCO.
With respect to the crowd test set ($\text{COCO}_{crd}$),
Table~\ref{tab:coco_all} shows that the performance is improved from
19.2\% to 20.5\% $\text{AP}$ for VGG-16. 
Since the larger number of categories in COCO makes distinguishing
instances harder, the improvement is smaller than the results on
$\text{VOC}_{crd}$. 
To demonstrate the general effectiveness of our IDNet, we also provide
the results when the backbone network is replaced by ResNet-101. 
The performance of IDNet is improved from 23.5\% $\text{AP}$ to
24.4\% $\text{AP}$ on the ResNet-101 backbone, compared with
Faster R-CNN, which shows the effectiveness of our IDNet on a stronger
backbone. 
We also observe that the improvement on the $\text{AP}_{75}$ is bigger than the improvement on the $\text{AP}_{50}$, which means the IDNet with the IDPP inference algorithm is effective for the localization accuracy.

For all COCO \texttt{val} images, the performance is
improved by 1.1\% $\text{AP}$ for the VGG-16 backbone and 1.2\%
$\text{AP}$ for the ResNet-101 backbone (Table~\ref{tab:coco_all}). 
We attribute the reason for the improvments to the fact that there are many similar categories in COCO, which has eight categories for each of 11 super categories on average.
Since a number of duplicated candidate boxes can be generated, our SS loss can remove duplicated bounding boxes to increase the final detection performance. 

\begin{figure}[!t]
\centering\includegraphics[width=\linewidth]{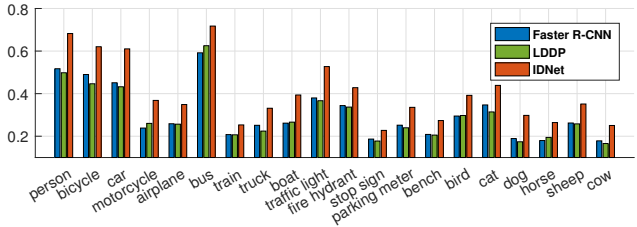}
\caption{\textbf{Probability of finding correct bounding boxes after training IDNet with SS loss.} For the evaluation, the IDNet is trained with COCO \texttt{train} set, and tested with COCO \texttt{val} set. The categories are sampled for the best view.}  \label{fig:coco_top5_prob} 
\end{figure}

To verify that SS loss affected the improvements, we extract candidate boxes having
detection scores over a fixed threshold (0.01) in Figure~\ref{fig:coco_top5_prob}.  
When a predicted box overlaps with the ground truth box by 0.5 of IoU
or more, we consider it as a correct box. 
We divide the number of correct boxes by the number of bounding boxes to check how many boxes are correctly classified.
Figure~\ref{fig:coco_top5_prob} shows that IDNet achieves 
superior performance on this measure for all categories compared to other methods.
On average, IDNet achieves 43.7\% while Faster R-CNN has 32.4\% and
LDDP has 32.9\% for COCO. 
The results indicate that the SS loss can successfully remove incorrectly classified bounding boxes.

\begin{table}[!t]
\centering
\caption{\textbf{Ablation study on COCO.} All results are from IDNet using VGG-16 as a backbone.}
\begin{tabular}{c|cc|cc} \thickhline
\multirow{2}{*}{Inference} & \multicolumn{2}{c|}{Loss} & \multicolumn{2}{c}{AP} \\ \cline{2-5}
 & SS & ID & $\text{COCO}_{all}$ & $\text{COCO}_{crd}$\\ \thickhline
NMS & & & 26.2 & 19.2 \\
NMS & \checkmark & & 27.0 & 19.6  \\
IDPP & & \checkmark & 26.5 & 19.7  \\
IDPP & \checkmark & \checkmark & \textbf{27.3} & \textbf{20.5} \\\hline
\thickhline 
\end{tabular} 
\label{tab:ablation}
\end{table}

\begin{figure}[!t]{\centering
\resizebox{\linewidth}{!}{
\begin{tabular}{cc}
\subfigure[A \textit{person} is not detected.]
{\centering	
  \includegraphics[width=0.5\linewidth]{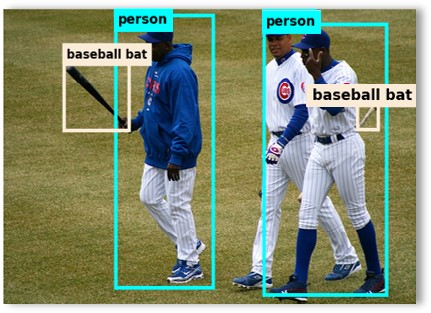}  
  \label{fig:a}
}%
\subfigure[A \textit{person} is detected.]
{\centering	
  \includegraphics[width=0.5\linewidth]{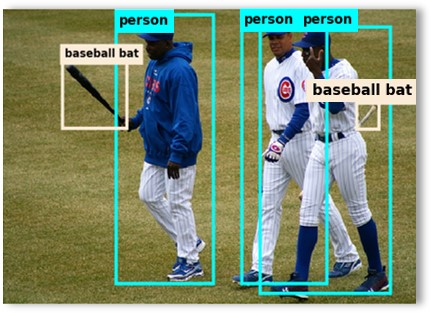}  
  \label{fig:a_idnet}
} \squeezeup \\ 
\subfigure[A \textit{sheep} is mistakenly detected.]
{\centering	
  \includegraphics[width=0.5\linewidth]{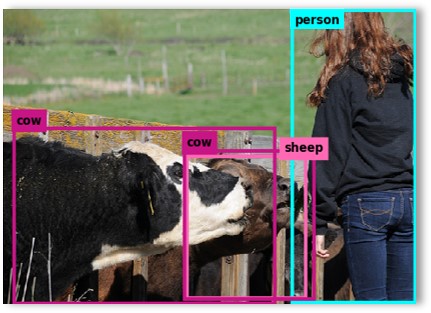} 
  \label{fig:b}
}%
\subfigure[A \textit{sheep} is removed.]
{\centering	
  \includegraphics[width=0.5\linewidth]{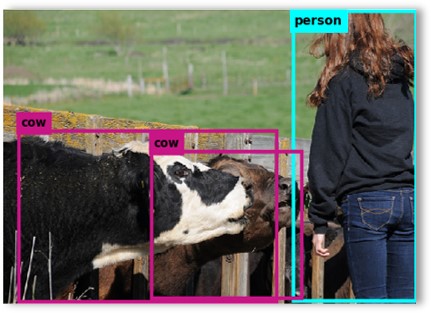} 
  \label{fig:b_idnet}
}%
\end{tabular} 
}
}
\caption{\textbf{Qualitative detection results of Faster R-CNN vs. IDNet.}
(a), (c) are results of Faster R-CNN and (b), (d) are results of IDNet. In (b), IDNet detect a \textit{person}, which is not detected on Faster R-CNN in (a). In (d), IDNet successfully suppresses an incorrect label, \textit{sheep}, while Faster R-CNN reports a \textit{sheep} in (c).}\label{fig:ablation} 
\end{figure} 

\begin{figure*}[!t]
\centering
  \includegraphics[width=\textwidth]{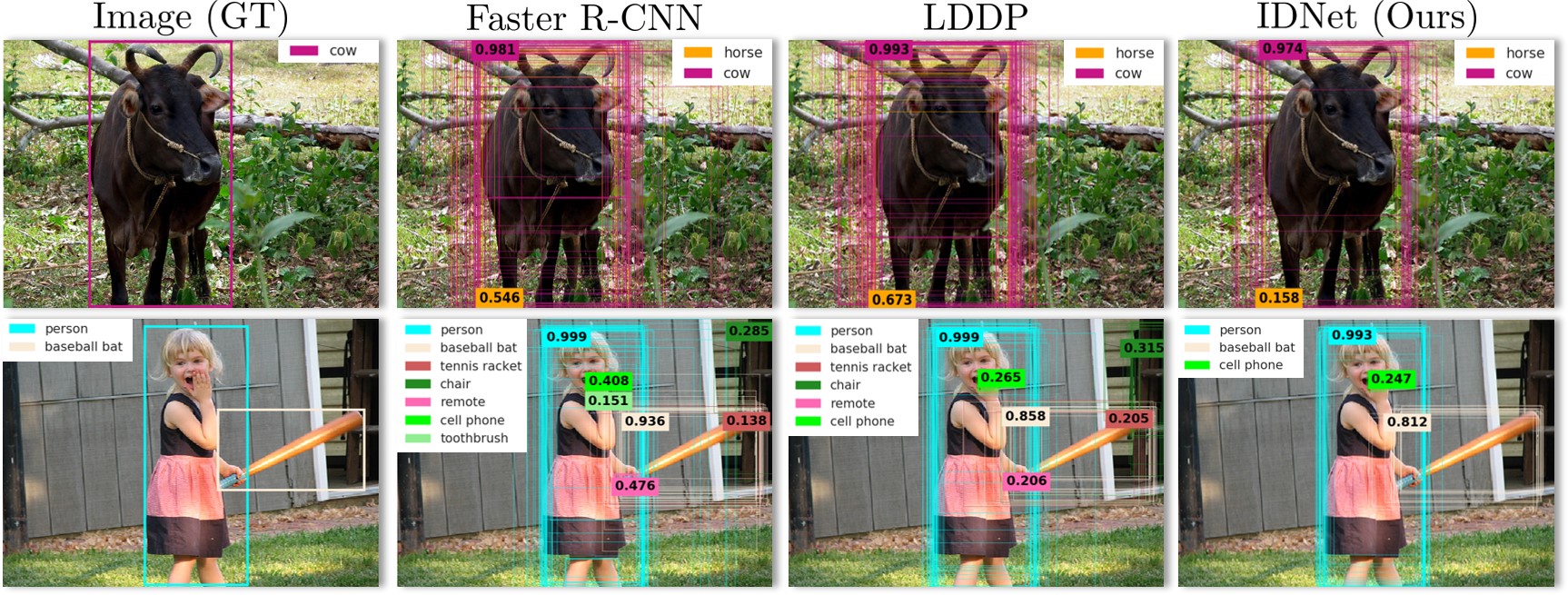}
  \caption{
  \textbf{Scores of candidate boxes after training with each method.} The leftmost column shows the ground truth boxes, and the other columns show the results of Faster R-CNN, LDDP, and IDNet from left to right. For each method, candidate boxes with scores over 0.1 and the maximum score of each category are visualized on each image. All methods are trained on COCO \texttt{train} set using VGG-16 as a backbone. Here, the IDNet only utilizes the SS loss.
} \label{fig:score_visu} 
\end{figure*}

\paragraph{Inference time.}
We measure the average inference time per image using VGG-16 as a backbone network on \texttt{minival} set of COCO, which is a subset of 5k samples from the \texttt{val} set. All running times are measured on a machine with Intel Core 3.7GHz CPU and Titan X GPU. 

Our algorithm takes 2.14 seconds to find candidate boxes and extract features of them, and 0.33 seconds to select bounding boxes using IDPP. Since Faster R-CNN takes 1.61 seconds and LDDP \cite{azadi2017learning} takes 1.55 seconds, an extra time of 0.86 seconds is needed for detecting objects in an image compared with Faster R-CNN and 0.92 seconds compared with LDDP. Although our algorithm takes more time to inference, it can be used in problems which require exact detections in a crowd.  

\subsection{Ablation Study} \label{sec:ablation}
We analyze the influence of the ID loss and SS loss in Table~\ref{tab:ablation}, where the IDNet is trained with COCO \texttt{train} set using VGG-16 as a backbone. 
In ablation studies, we check our IDNet with two post-processing
methods: NMS and IDPP.
In the first two rows in Table~\ref{tab:ablation}, we use NMS for the experiments that do not use the ID loss, since IDPP uses the trained features with the ID loss. 
In the last two rows of Table~\ref{tab:ablation}, we use IDPP with a trained RIN module.

\paragraph{Instance-aware identity loss.} 
The ID loss is made to be effective for detecting objects in a crowded
scene. 
In the third row of Table~\ref{tab:ablation}, the performance is
improved from 19.2\% to 19.7\% AP on $\text{COCO}_{crd}$. 
Comparing the second row and the last row, the performance is improved by 0.9\% AP.
In Figure~\ref{fig:a}, a \textit{person} is not detected in
Faster R-CNN, while our IDNet detects the \textit{person} in Figure~\ref{fig:a_idnet} since IDNet
learns to discriminate different objects. 
This result indicates that the ID loss is effective for detecting objects in proximity.

\paragraph{Sparse-score loss.} 

Since the SS loss is designed to remove incorrectly classified bounding boxes, the SS loss is effective for all testing images.
Thus, we focus on the results on $\text{COCO}_{all}$ column in Table~\ref{tab:ablation}. 
The results show that as
the SS loss is used, the performance is improved by 0.8\% AP.

In Figure~\ref{fig:b}, a \textit{sheep} is erroneously detected
for a \textit{cow}, while our IDNet removes this erroneous detection
of a \textit{sheep} in Figure~\ref{fig:b_idnet} as IDNet learns to remove incorrectly classified bounding boxes. 
It shows that the SS loss can alleviate duplicated bounding box
problem in a detector.  

Since Figure~\ref{fig:ablation} only shows the final detections, we
visualize images with candidate boxes in Figure~\ref{fig:score_visu}
to show the changes in detection scores. 
The score threshold is fixed to 0.1 and the highest score in each
category is written in each image. 

We first compare the result with Faster R-CNN. Since Faster R-CNN does not have any loss to decrease the scores of incorrect categories, the highest score of a \textit{horse} in Faster R-CNN is 0.546 while the score in IDNet is 0.158 (see the first row of Figure~\ref{fig:score_visu}).
For images in the second row of Figure~\ref{fig:score_visu}, the maximum score of an incorrect category, \textit{remote}, is 0.476 in Faster R-CNN, while the maximum score of a \textit{remote} is under the threshold (0.1) in IDNet. 

We also compare the result with LDDP~\cite{azadi2017learning}. The LDDP loss~\cite{azadi2017learning} is defined to increase the score of a single subset using a category-level relationship, while our SS loss is defined to decrease scores of all possible subsets containing incorrect candidate boxes using an instance-level relationship between candidate boxes.
Thus, after softmax is applied to scores, the SS loss can better suppress the detection scores of bounding boxes with incorrect categories. 
For example, as shown in the third and last columns of
Figure~\ref{fig:score_visu}, given a \textit{cow} image, the detection
score for a \textit{horse} is decreased from 0.673 (LDDP) to 0.158
(IDNet). 
It shows that the SS loss can successfully suppress scores of
duplicated bounding boxes around a correct bounding box as expected. 

\section{Conclusion}

We propose IDNet which tackles two challenges in object
detection by introducing two novel losses. 
First, we propose the ID loss for detecting overlapped objects.
Second, the SS loss is introduced to suppress erroneous detections of
wrong categories. 
By introducing these two losses using DPPs,
we demonstrate that learning instance-level relationship is useful for accurate detection.
IDNet performs favorably for overall test sets and achieves
significant improvements on the crowd sets.
Additionally, the ablation studies show that IDNet
learns to suppress erroneous detections of wrong categories.
While the inference time is moderately slower than other detection
methods, our algorithm is useful for real-world situations which
require separating objects in proximity.

\clearpage
{\LARGE{\textbf{Appendix}}}
\appendix

\begin{table*}[!t]
  \centering
  \caption{
  \textbf{Notations in this paper.}
  }
  \begin{tabular}{l|l|l}
    \thickhline
    \multicolumn{1}{l}{Notation} & \multicolumn{1}{l}{Definition} & \multicolumn{1}{l}{Description} \\
    \thickhline
    RoIs &
    - &
    Region of interest boxes which are proposed from RPN. \\
    $\textbf{b}$ &
    - &
    Candidate bounding boxes which are proposed from RCN. \\
    ${\IoU}_{ij}$ &
    ${\#(\textbf{b}_i \cap \textbf{b}_j)}/{\#(\textbf{b}_i \cup \textbf{b}_j)}$ &
    Intersection over union (IoU) of two bounding boxes. \\
    $\textbf{q}$ &
    - &
    Detection score corresponding to the candidate bounding boxes. \\
    $\matr{V}_{i}$ &
    ${\textbf{F}_{i}}/{{\|\textbf{F}_{i}\|}_2}$ &
    Normalized feature of a bounding box $i$. \\
    $\matr{S}_{ij}$ &
    $\lambda \cdot \matr{V}_{i}\matr{V}_{j}^T + (1 - \lambda) \cdot \IoU_{ij}$ &
    Similarity between box $i$ and $j$. $0 < \lambda < 1$.\\
    $\matr{L}$ &
    $\matr{S} \odot \textbf{q}\textbf{q}^T$ &
    Kernel matrix of DPPs. \\
    \thickhline
  \end{tabular}
  \label{tab:notation}
\end{table*}

\section{Notations} \label{sec:notations}
In Table~\ref{tab:notation}, notations used in this paper are described.

\section{Gradient of losses} \label{sec:gradient}
In this section, we derive the gradients of the proposed instance-aware detection loss (ID loss) and sparse-score loss (SS loss). For notational convenience, we assume that the matrix $\matr{M}_{x}$ has the same dimension as $\matr{M}$ and its entries corresponding to $x$ is copied from $\matr{M}$ while remaining entries are filled with zero, for any matrix $\matr{M}$ and indices $x$. 

\subsection{Gradient of Instance-Aware Detection Loss}
Here, we show the gradient with respect to the normalized feature ($\matr{V}$). 
As the derivative of the log-determinant is $\partial\log\det(\matr{L}) = \partial(\Tr(\log(\matr{L}))) = \Tr(\matr{L}^{-T}\partial\matr{L})$, the derivative of intra-class ID loss is as follows:
\begin{equation}
\setlength{\jot}{5pt}
\begin{aligned}
& \{ \partial\mathcal{L}^{ic}_{ID}({Y}_{C_k}, \mathcal{Y}_{C_k}) \}_{C_k} \\
&= - \partial\log \det (\matr{L}_{{Y}_{C_k}}) + \partial\log \det (\matr{L}_{\mathcal{Y}_{C_k}}+\matr{I}_{\mathcal{Y}_{C_k}})\\
&= - \partial\Tr (\log(\matr{L}_{{Y}_{C_k}}))+\partial\Tr (\log (\matr{L}_{\mathcal{Y}_{C_k}}+\matr{I}_{\mathcal{Y}_{C_k}})) \\
&= - \Tr (\matr{L}_{{Y}_{C_k}}^{-T}\partial\matr{L}_{{Y}_{C_k}}) \\
 & \;\;\;\; + \Tr ((\matr{L}_{\mathcal{Y}_{C_k}}+\matr{I}_{\mathcal{Y}_{C_k}})^{-T}\partial(\matr{L}_{\mathcal{Y}_{C_k}}+\matr{I}_{\mathcal{Y}_{C_k}})) \\
&= -\langle {\matr{L}_{{Y}_{C_k}}}^{-1}, \; \partial\matr{L}_{{Y}_{C_k}} \rangle \\
& \;\;\;\; + \langle {(\matr{L}_{\mathcal{Y}_{C_k}}+\matr{I}_{\mathcal{Y}_{C_k}})}^{-1},\; \partial\matr{L}_{\mathcal{Y}_{C_k}} \rangle ,\\
\end{aligned}
\end{equation}
where $\langle\cdot,\cdot\rangle$ is the Frobenius inner product, $\odot$ is the element-wise multiplication, and $k \in \{1,..., n_c\}$ is the $k$th category. Note that the $n_c$ is the number of categories. We only calculate the gradient of the ID loss on the similarity feature ($\matr{V}$), where $\matr{S} = \lambda \cdot \matr{V}\matr{V}^T + (1-\lambda) \cdot \IoU$.  Since $\IoU$ is a constant, the derivative of $\matr{L}$ is as follows:
\begin{equation}
\begin{aligned}
\partial\matr{L} = \lambda \cdot \matr{Q} \odot (\partial\matr{V}\matr{V}^T + \matr{V}\partial\matr{V}^T),
\end{aligned}
\end{equation}
where $\matr{Q} = \textbf{q}\textbf{q}^T$. Note that $\matr{Q}$ is fixed while deriving gradient of the ID loss. Using the property that $\langle \matr{A} ,\; \matr{B} \odot \matr{C} \rangle = \langle \matr{A} \odot \matr{B} ,\; \matr{C} \rangle$, where $\matr{A}, \matr{B}, \text{ and } \matr{C}$ are arbitrary matrices, we can derive this:
\begin{equation}
\begin{aligned}
\{& \partial\mathcal{L}^{ic}_{ID}({Y}_{C_k}, \mathcal{Y}_{C_k}) \}_{C_k} \\
&= -2\lambda \cdot \langle {(\matr{Q}_{{Y}_{C_k}} \odot \matr{L}_{{Y}_{C_k}})}^{-1}{\matr{V}_{{Y}_{C_k}},\; \partial\matr{V}_{{Y}_{C_k}}\rangle}\\
&\hspace{3ex}{+2\lambda \cdot \langle \matr{Q}_{\mathcal{Y}_{C_k}} \odot{( \matr{L}_{\mathcal{Y}_{C_k}}+\matr{I}_{\mathcal{Y}_{C_k}})}^{-1}\matr{V}_{\mathcal{Y}_{C_k}},\; \partial\matr{V}_{\mathcal{Y}_{C_k}}\rangle}.\\
\end{aligned}
\end{equation}
By seeing the matrix in element-wise,
\begin{equation}
\begin{aligned} \label{eq:ic_eq}
\Bigg\{&\frac{\partial\mathcal{L}^{ic}_{ID}({Y}_{C_k}, \mathcal{Y}_{C_k})}{\partial\matr{V}} \Bigg\}_{C_k} \\ &= -2\lambda \cdot (\matr{Q}_{{Y}_{C_k}} \odot \matr{L}_{{Y}_{C_k}})^{-1}\matr{V}_{{Y}_{C_k}}\\
& \hspace{3ex} + 2\lambda \cdot \matr{Q}_{\mathcal{Y}_{C_k}} \odot (\matr{L}_{\mathcal{Y}_{C_k}} + \matr{I}_{\mathcal{Y}_{C_k}})^{-1} \matr{V}_{\mathcal{Y}_{C_k}}.
\end{aligned}
\end{equation}

Since the gradient of $\mathcal{L}^{all}_{ID}$ is similar with a gradient of $\mathcal{L}^{ic}_{ID}$, we omit the derivation of that. 
Then, we can construct the gradient of ID loss by summing up (\ref{eq:ic_eq}) for all batches and categories:
\begin{equation}
\begin{aligned}
 &\frac{\partial\mathcal{L}_{ID}(Y_{rep}, \mathcal{Y}_{s}, {Y}_{C_k}, \mathcal{Y}_{C_k} )}{\partial\matr{V}} \\=&\frac{\partial\mathcal{L}^{all}_{ID}(Y_{rep}, \mathcal{Y}_{s})}{\partial\matr{V}} + \sum_{k=1}^{n_c} \Bigg\{ \frac{\partial\mathcal{L}^{ic}_{ID}({Y}_{C_k}, \mathcal{Y}_{C_k})}{\partial\matr{V}} \Bigg\}_{C_k}.
\end{aligned}
\end{equation}
\subsection{Gradient of Sparse-Score Loss}
The derivation for calculating the gradient of the SS loss is similar with the derivation of the instance-aware detection loss, while the gradient for the SS loss is derived over the quality ($\textbf{q}$). Note that $\matr{S}$ is fixed while deriving gradient of the SS loss. The derivative of the SS loss is as follows:
\begin{equation}
\setlength{\jot}{5pt}
\begin{aligned}
& \partial\mathcal{L}_{SS}({Y}_{pos}, \mathcal{Y}_{m}) \\
&= - \partial\log \det (\matr{L}_{{Y}_{pos}} + \matr{I}_{{Y}_{pos}}) + \partial\log \det (\matr{L}_{\mathcal{Y}_{m}}+\matr{I}_{\mathcal{Y}_{m}})\\
&= - \partial \Tr (\log(\matr{L}_{{Y}_{pos}} + \matr{I}_{{Y}_{pos}}))+\partial \Tr (\log (\matr{L}_{\mathcal{Y}_{m}}+\matr{I}_{\mathcal{Y}_{m}})) \\
&= - \Tr ((\matr{L}_{{Y}_{pos}} + \matr{I}_{{Y}_{pos}})^{-T}\partial\matr{L}_{{Y}_{pos}} + \matr{I}_{{Y}_{pos}}) \\ & \;\;\;\; + \Tr ((\matr{L}_{\mathcal{Y}_{m}}+\matr{I}_{\mathcal{Y}_{m}})^{-T}\partial(\matr{L}_{\mathcal{Y}_{m}}+\matr{I}_{\mathcal{Y}_{m}})) \\
&= -\langle {(\matr{L}_{{Y}_{pos}} + \matr{I}_{{Y}_{pos}}})^{-1}, \; \partial\matr{L}_{{Y}_{pos}} + \matr{I}_{{Y}_{pos}} \rangle \\
& \;\;\;\; + \langle {(\matr{L}_{\mathcal{Y}_{m}}+\matr{I}_{\mathcal{Y}_{m}})}^{-1},\; \partial\matr{L}_{\mathcal{Y}_{m}} \rangle.\\
\end{aligned}
\end{equation}
Similar to the derivation of ID loss, by using the following properties,
\begin{equation}
\begin{aligned}
&\partial\matr{L} =  \matr{S} \odot (\partial\matr{q}\matr{q}^T + \matr{q}\partial\matr{q}^T),\\
&\langle \matr{A} ,\; \matr{B} \odot \matr{C} \rangle = \langle \matr{A} \odot \matr{B} ,\; \matr{C} \rangle,
\end{aligned}
\end{equation}
we can derive this:
\begin{equation}
\begin{aligned}
& \partial\mathcal{L}_{SS}({Y}_{pos}, \mathcal{Y}_{m}) \\
&= -2 \cdot \langle {\matr{S}_{{Y}_{pos}} \odot (\matr{L}_{{Y}_{pos}}+\matr{I}_{{Y}_{pos}})^{-1}}{\textbf{q}_{{Y}_{pos}},\; \partial\textbf{q}_{{Y}_{pos}}\rangle}\\
&\hspace{3ex}{+2 \cdot \langle \matr{S}_{\mathcal{Y}_{m}} \odot{( \matr{L}_{\mathcal{Y}_{m}}+\matr{I}_{\mathcal{Y}_{m}})}^{-1}\textbf{q}_{\mathcal{Y}_{m}},\;
{\partial\textbf{q}_{\mathcal{Y}_{m}}\rangle}.}
\end{aligned}
\end{equation}
Thus, the final derivative of SS loss is as follows:
\begin{equation}
\begin{aligned}
& \frac{\partial\mathcal{L}_{SS}({Y}_{pos}, \mathcal{Y}_{m})}{\partial\textbf{q}} \\
&= -2 \cdot \matr{S}_{{Y}_{pos}} \odot (\matr{L}_{{Y}_{pos}} + \matr{I}_{{Y}_{pos}})^{-1}\textbf{q}_{{Y}_{pos}}\\
& \hspace{2.5ex} + 2 \cdot \matr{S}_{\mathcal{Y}_{m}} \odot (\matr{L}_{\mathcal{Y}_{m}} + \matr{I}_{\mathcal{Y}_{m}})^{-1} \textbf{q}_{\mathcal{Y}_{m}}.
\end{aligned}
\end{equation}

\section{Network Architecture} \label{sec:net_archi}
As shown in Table~\ref{tab:rin}. the RIN consists of seven convolutional layers, three fully connected layers, three max-pooling layers, and one crop and resize layer. Since RIN utilizes parameters of a backbone network, the size of input channel ($c_{in}$) is chosen according to the backbone network, e.g, 64 for VGG-16 and ResNet-101. The parameters $c_1 = 64, c_2 = 128,\text{ and } c_3 = 128$ are used for training with VOC. For COCO,  $c_1 = 128, c_2 = 256,\text{ and } c_3 = 256$ are used. At the end of each convolutional and fully-connected layer except the last layer has a batch normalization \cite{ioffe2015batch} and a rectified linear unit (ReLU) in order. We set all convolutional layers to have filters with a size of 3 $\times$ 3 pixels and a stride of one. 

\begin{table}[!h]
\centering
\caption{\textbf{RIN architecture.}}
\label{tab:rin}
\resizebox{\linewidth}{!}{
\begin{tabular}{cccc}
\thickhline
\multicolumn{1}{c}{Layer} & \multicolumn{1}{c}{Type} & \multicolumn{1}{c}{Parameter} & \multicolumn{1}{c}{Remark} \\ 
\thickhline
0 & Convolution & $c_{in}\times\text{3}\times\text{3}\times c_1$ & stride 1 \\
1 & Convolution & $c_1\times\text{3}\times\text{3}\times c_1$ & stride 1 \\
2 & Convolution & $c_1\times\text{3}\times\text{3}\times c_2$ & stride 1 \\
3 & Convolution & $c_2\times\text{3}\times\text{3}\times c_2$ & stride 1 \\
4 & Max pooling & - & size 2$\times$2, stride 2 \\
5 & Convolution & $c_2\times\text{3}\times\text{3}\times c_3$ & stride 1 \\
6 & Convolution & $c_3\times\text{3}\times\text{3}\times c_3$ & stride 1 \\
7 & Convolution & $c_3\times\text{3}\times\text{3}\times c_3$ & stride 1 \\
8 & Crop and resize & - & size 15$\times$15 \\
9 & Fully connected & ($15^2$ $\cdot$ $c_3$)$\times$1000 & - \\
10 & Fully connected & 1000x1000 & - \\
11 & Fully connected & 1000x256 & - \\ \thickhline
\end{tabular}}
\end{table}

\section{More Experimental Results}
In this section, we provide full results on PASCAL VOC and MS COCO datasets. For the results on all test images are in the Table~\ref{tab:pascal0712_orig} and Table~\ref{tab:coco_orig}. Table~\ref{tab:pascal0712_orig_overlap} and Table~\ref{tab:coco_orig_overlap} show the results on the crowd sets.

\begin{figure}[!t]
\centering\includegraphics[width=\linewidth]{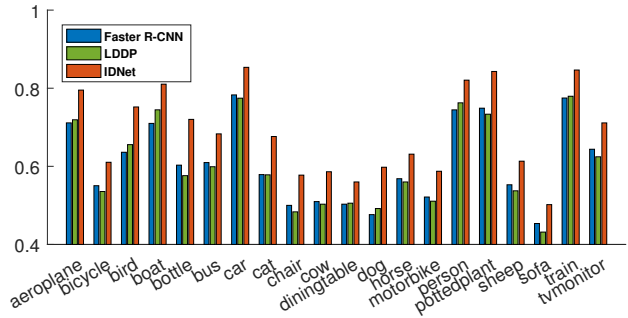}
\caption{\textbf{Probability of finding correct bounding boxes after training IDNet with SS loss.} For the evaluation, the IDNet is trained with VOC. The categories are sampled for the best view.}  \label{fig:pascal07_overlap} 
\end{figure}

\begin{figure}[!t]{
\centering\resizebox{\linewidth}{!}{\centering\includegraphics[width=\linewidth]{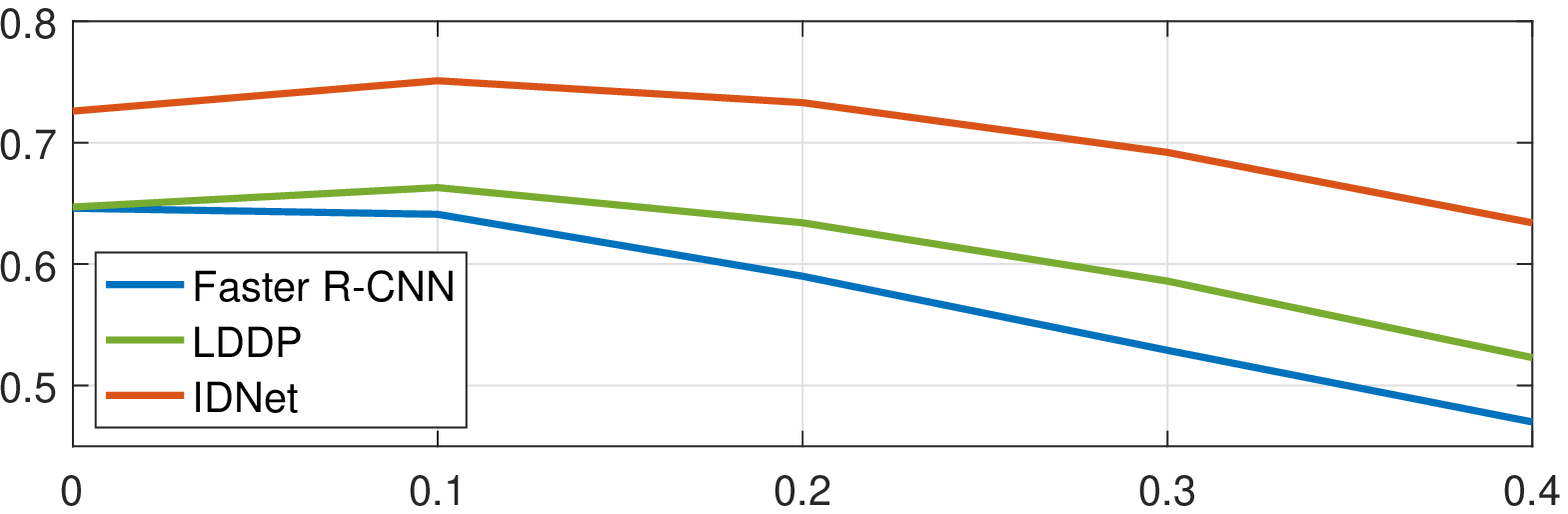}}
\caption{\textbf{Recall curves of Faster R-CNN, LDDP, and IDNet on COCO.} The results are evaluated at different overlap IoU thresholds, from .0 to .4. Our proposed IDNet has a higher crowd recall and effectively detects object with high overlaps.}
\label{fig:recall_coco} } \expandup
\end{figure}

Additionally, Figure~\ref{fig:pascal07_overlap} is the graph showing the impact of SS loss on VOC dataset and Figure~\ref{fig:recall_coco} shows the recall graph for COCO dataset.

\section{Example Visualization} \label{sec:visualize}
We visualize qualitative results of IDNet on VOC 2007 and MS COCO. For comparison, we also visualize the ground truth bounding boxes in each image, and the results of Faster R-CNN and LDDP. For Faster R-CNN and LDDP, only bounding boxes with a score threshold of 0.6 are visualized. The threshold is designated in their paper \cite{azadi2017learning}. For IDNet, we use 0.2 as a score threshold.

\begin{table*}[!t]
\centering
\caption{\textbf{Detection results on VOC 2007 \texttt{test} set.} Legend: \textbf{07:} VOC 2007 \texttt{trainval} set, \textbf{07+12:} VOC 0712 \texttt{trainval} set. All methods are trained with the multi-task loss, using a VGG-16 backbone network.}
\resizebox{\textwidth}{!}{\begin{tabular}{ccc|ccc|c|cccccccccccccccccccc}
 \thickhline
Method & Inference & Train & mAP  & aero & bike & bird & boat & bottle & bus  & car  & cat  & chair & cow  & table & dog  & horse & mbike & person & plant & sheep & sofa & train & tv   \\ \thickhline
Faster R-CNN~\cite{ren2015faster} & NMS & 07 & 71.4 & 70.4 & 78.2 & 69.7 & \textbf{58.9} & \textbf{56.9} & 79.5 & 83.0 & 84.3 & 53.3 & 78.6 & 64.5 & 81.7 & 83.7 & 76.1 & 77.9 & 45.4 & 70.5 & 66.7 & 74.3 & 73.3 \\
LDDP~\cite{azadi2017learning} & LDPP & 07 & 70.9 & 67.7 & 79.2 & 68.2 & 57.9 & 53.9 & 75.2 & 7979 & 84.8 & 53.7 & 79.2 & 67.5 & 80.9 & 84.0 & 75.7 & 78.0 & 44.7 & 73.3 & 66.7 & 73.8 & 73.1 \\
$\text{IDNet}^*$ & IDPP & 07 & 71.9 & 71.3 & 79.1 & 70.8 & 57.9 & 53.1 & 77.5& 84.2 & 85.8 & 53.0 & 80.4 & 69.1 & 80.7 & 84.3 &75.8 & 79.6 & 44.0 & 75.1 & 66.8 & 76.5 & 73.2 \\ \hline
Faster R-CNN~\cite{ren2015faster} & NMS  & 07+12 & 75.8 & 77.2 & \textbf{84.1} & 74.8 & \textbf{67.3} & 65.5 & 82.0 & 87.4 & \textbf{87.9} & 58.7 & 81.5 & 69.8 & 85.0 & 85.1 & 77.7 & 79.2 & 47.2 & 75.4 & 71.8 & 82.3 & 75.8 \\
LDDP~\cite{azadi2017learning} & LDPP & 07+12& 76.4 & 76.9 & 83.0 & 75.0 & 66.5 & 64.3 & 83.4 & 87.5 & 87.7 & 61.2 & 81.5 & 70.0 & 86.0 & 84.9 & \textbf{81.9} & \textbf{83.3} & \textbf{48.6} & 75.7 & 72.3 & 82.6 & 76.5 \\
$\text{IDNet}^*$ & IDPP & 07+12 &76.6 & 78.8 & 82.8 & 75.9 & 66.3 & 66.6 & 82.9 & 88.1 & 87.2 & 59.6 & 82.4 & 70.6 & 85.1 & 85.7 & 80.7 & 82.6 & 50.0 & 78.3 & 70.9 & 82.8 & 75.5 \\
\thickhline
\end{tabular}}
\label{tab:pascal0712_orig} \expandup \expandup
\end{table*}

\begin{table*}[!t]
\centering
\caption{\textbf{Detection results on VOC 2007 crowd set.} Legend: \textbf{07:} VOC 2007 \texttt{trainval} set, \textbf{07+12:} VOC 0712 \texttt{trainval} set. All methods are trained with the multi-task loss, using a VGG-16 backbone network.}
\resizebox{\textwidth}{!}{\begin{tabular}{ccc|c|cccccccccccccccccccc}
 \thickhline
Method & Inference & Train & mAP & aero & bike & bird & boat & bottle & bus  & car  & cat  & chair & cow  & table & dog  & horse & mbike & person & plant & sheep & sofa & train & tv   \\ \thickhline
Faster R-CNN~\cite{ren2015faster} & NMS & 07 & 56.0 & 45.5 & 56.0 & 44.2 & 42.0 & 57.4 & 54.5 & 70.3 & 37.4 & 47.2 & 67.8 & 65.4 & 56.4 & 63.0 & 61.4 & 67.8 & 30.0 & 66.4 & 53.4 & 63.6 & \textbf{70.6} \\
LDDP~\cite{azadi2017learning} & LDPP & 07 & 57.7 & 38.2 & \textbf{61.4} & 47.9 & 37.7 & 54.3 & 54.5 & 74.6 & \textbf{48.1} & 49.5 & \textbf{76.1} & \textbf{70.3} & 60.3 & 63.3 & 60.3 & 73.7 & 31.4 & 70.5 & 52.3 & 63.6 & 66.3 \\
$\text{IDNet}^*$ & IDPP & 07 & \textbf{61.8} & \textbf{65.5} & 59.6 & \textbf{56.2} & \textbf{49.8} &  \textbf{60.2} &  \textbf{61.2} & 76.0 & 38.7 & 50.1 & 67.4 & 65.5 & \textbf{68.0} & 67.8 & \textbf{64.2} & \textbf{74.0} & \textbf{35.8} & \textbf{75.6} & 50.6 & \textbf{81.8} & 68.6 \\ 
 \hline
Faster R-CNN~\cite{ren2015faster} & NMS  & 07+12 &62.0 & \textbf{100.0} & 59.4 & \textbf{60.1} & 28.5 & 61.3 &53.2 & 72.0 & 51.4 & 51.9 & 67.0 & 67.0 & 55.1 & 76.9 & 71.4 & 69.4 & 32.6 & 67.5 & 61.1 & \textbf{63.6} & 70.2 \\
LDDP~\cite{azadi2017learning} & LDPP & 07+12 &  63.1 & 78.5 & 64.6 & 55.6 & \textbf{34.8} & 60.3 & 52.1 & 76.9 & \textbf{55.4} & \textbf{56.7} & 72.8 & \textbf{69.0} & \textbf{69.0} & 73.2 & 69.3 & \textbf{76.3} & \textbf{41.4} & 73.4 & 48.2 & \textbf{63.6} & \textbf{70.5} \\
$\text{IDNet}^*$ & IDPP & 07+12 & \textbf{64.5} & 88.3 & \textbf{68.8} & 59.8 & 31.9 & \textbf{64.1} &\textbf{61.7} & \textbf{79.0} & 48.7 & 54.4 & 72.3 & 66.5 & 64.2 & \textbf{77.7} & \textbf{71.7} & 75.6 & 37.7 & \textbf{77.0} & 57.5 & \textbf{63.6} & 70.0 \\
\thickhline
\end{tabular}} 
\label{tab:pascal0712_orig_overlap} \expandup \expandup
\end{table*}

\begin{table*}[!t]
\centering
\caption{\textbf{Detection results on MS COCO \texttt{val} set.} All methods are trained on MS COCO \texttt{train} set with the multi-task loss.}
\begin{center}
\resizebox{\textwidth}{!}{\begin{tabular}{ccc|ccc|ccc|ccc|ccc}
\thickhline  
Method & Inference & Backbone & AP & $\text{AP}_{50}$ & $\text{AP}_{75}$ & $\text{AP}_{S}$ & $\text{AP}_{M}$ & $\text{AP}_{L}$ & $\text{AR}_\text{1}$ & $\text{AR}_\text{10}$ & $\text{AR}_\text{100}$ & $\text{AR}_{S}$ & $\text{AR}_{M}$ & $\text{AR}_{L}$ \\ \thickhline
Faster R-CNN~\cite{ren2015faster} & NMS & VGG-16 & 26.2 & 46.6 & 26.9 & 10.3 & 29.3 & 36.4 & 25.5 & 38.1 & 39.0 & 17.9 & 44.0 & 55.7 \\
LDDP~\cite{azadi2017learning} & LDPP & VGG-16 & 26.4 & 46.7 & 26.8 & 10.5 & 29.4 & 36.8 & 25.0 & 37.4 & 38.4 & 16.0 & 43.1 & 55.3 \\ 
IDNet & IDPP & VGG-16 & \textbf{27.3} & \textbf{47.6} & \textbf{28.2} & \textbf{10.9} & \textbf{30.1} & \textbf{38.0} & \textbf{25.9} & \textbf{39.4} & \textbf{40.6} & \textbf{18.6} & \textbf{45.1} & \textbf{58.9} \\ \hline
Faster R-CNN~\cite{ren2015faster} & NMS & ResNet-101 & 31.5 & 52.0 & 33.5 & 12.5 & 35.2 & 45.9 & 29.2 & 43.2 & 44.2 & 20.6 & 49.9 & 63.8 \\
LDDP~\cite{azadi2017learning} & LDPP & ResNet-101 & 31.4 & 51.7 & 33.4 & 12.3 & 35.3 & 46.0 & 28.5 & 41.9 & 42.9 & 18.2 & 48.2 & 63.4 \\ 
IDNet & IDPP & ResNet-101 & \textbf{32.7} & \textbf{53.1} & \textbf{34.8} & \textbf{13.1} & \textbf{36.4} & \textbf{47.6} & \textbf{29.5} & \textbf{44.3} & \textbf{45.6} & \textbf{21.2} & \textbf{51.2} & \textbf{65.8} \\ \thickhline
\end{tabular}} 
\end{center}
\label{tab:coco_orig} \expandup \expandup
\end{table*}

\begin{table*}[!t]
\centering
\caption{\textbf{Detection results on MS COCO crowd set.} All methods are trained on MS COCO \texttt{train} set with the multi-task loss.}
\begin{center}
\resizebox{\textwidth}{!}{\begin{tabular}{ccc|ccc|ccc|ccc|ccc}
\thickhline  
Method & Inference & Backbone & AP & $\text{AP}_{50}$ & $\text{AP}_{75}$ & $\text{AP}_{S}$ & $\text{AP}_{M}$ & $\text{AP}_{L}$ & $\text{AR}_\text{1}$ & $\text{AR}_\text{10}$ & $\text{AR}_\text{100}$ & $\text{AR}_{S}$ & $\text{AR}_{M}$ & $\text{AR}_{L}$ \\ \thickhline
Faster R-CNN~\cite{ren2015faster} & NMS & VGG-16 & 19.2 & 36.9 & 18.4 & 8.5 & 24.3 & 31.0 & 17.0 & 28.6 & 29.6 & 13.4 & 36.4 & 47.8 \\
LDDP~\cite{azadi2017learning} & LDPP & VGG-16 & 19.6 & 37.9 & 18.6 & 8.9 & 24.6 & 31.6 & 16.6 & 28.4 & 29.6 & 12.9 & 36.4 & 47.7 \\ 
IDNet & IDPP & VGG-16 & \textbf{20.5} & \textbf{38.2} & \textbf{20.0} & \textbf{9.1} & \textbf{25.7} & \textbf{33.0} & 17.0 & \textbf{30.9} & \textbf{33.2} & \textbf{14.4} & \textbf{39.2} & \textbf{56.0} \\ \hline
Faster R-CNN~\cite{ren2015faster} & NMS & ResNet-101 &23.5 & 42.5 & 23.0 & 10.4 & 29.6 & 38.5 & 19.3 & 32.8 & 34.0 & 16.1 & 41.7 & 54.6 \\
LDDP~\cite{azadi2017learning} & LDPP & ResNet-101 & 23.8 & 43.0 & 23.4 & 10.5 & 30.0 & 39.4 & 19.2 & 32.4 & 33.7 & 15.0 & 41.4 & 55.2 \\ 
IDNet & IDPP & ResNet-101 & \textbf{24.4} & \textbf{43.4} & \textbf{24.4} & \textbf{10.9} & \textbf{30.6} & \textbf{40.0} & \textbf{19.6} & \textbf{33.7} & \textbf{34.8} & \textbf{16.5} & \textbf{42.4} & \textbf{56.4} \\ \thickhline
\end{tabular}} 
\end{center}
\label{tab:coco_orig_overlap} \expandup \expandup
\end{table*}

\begin{figure}[!t]
\centering
\subfigure{
  \centering
  \includegraphics[width=0.47\linewidth]{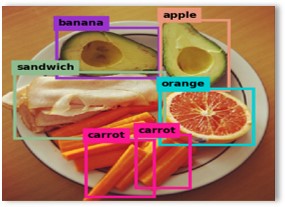}}
\subfigure{
  \centering
  \includegraphics[width=0.47\linewidth]{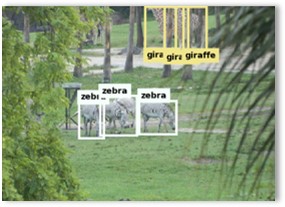}}
  \caption{
\textbf{Failure cases of IDNet.} \textbf{Top:} A detector find an incorrect category; \textbf{Bottom:} A detector cannot distinguish a completely occluded object. The class labels are arranged for the best view.
} \label{fig:fail} \expandup
\end{figure}

\paragraph{Failure cases analysis.} \label{subsec:failure}
The top image of Figure~\ref{fig:fail} shows that the detector detected the bounding box of the wrong category for avocados. This means that the detector has found a class similar to avocado, such as banana and apple because there are no categories in a dataset. This case suggests that there is a need to suppress further scores for pictures in the absence of a detection class, i.e., background category. In the bottom of Figure~\ref{fig:fail}, a giraffe is hidden behind two trees. If there is an occlusion for an object, detectors tend not to notice that it is a single object. Then detectors choose several bounding boxes for the object. Since IDPP tries to find the most representative bounding boxes, it would select all of the created bounding boxes, which increases the number of false detections.

\begin{figure*}[!ht]
\begin{center}
    {\includegraphics[width=\linewidth]{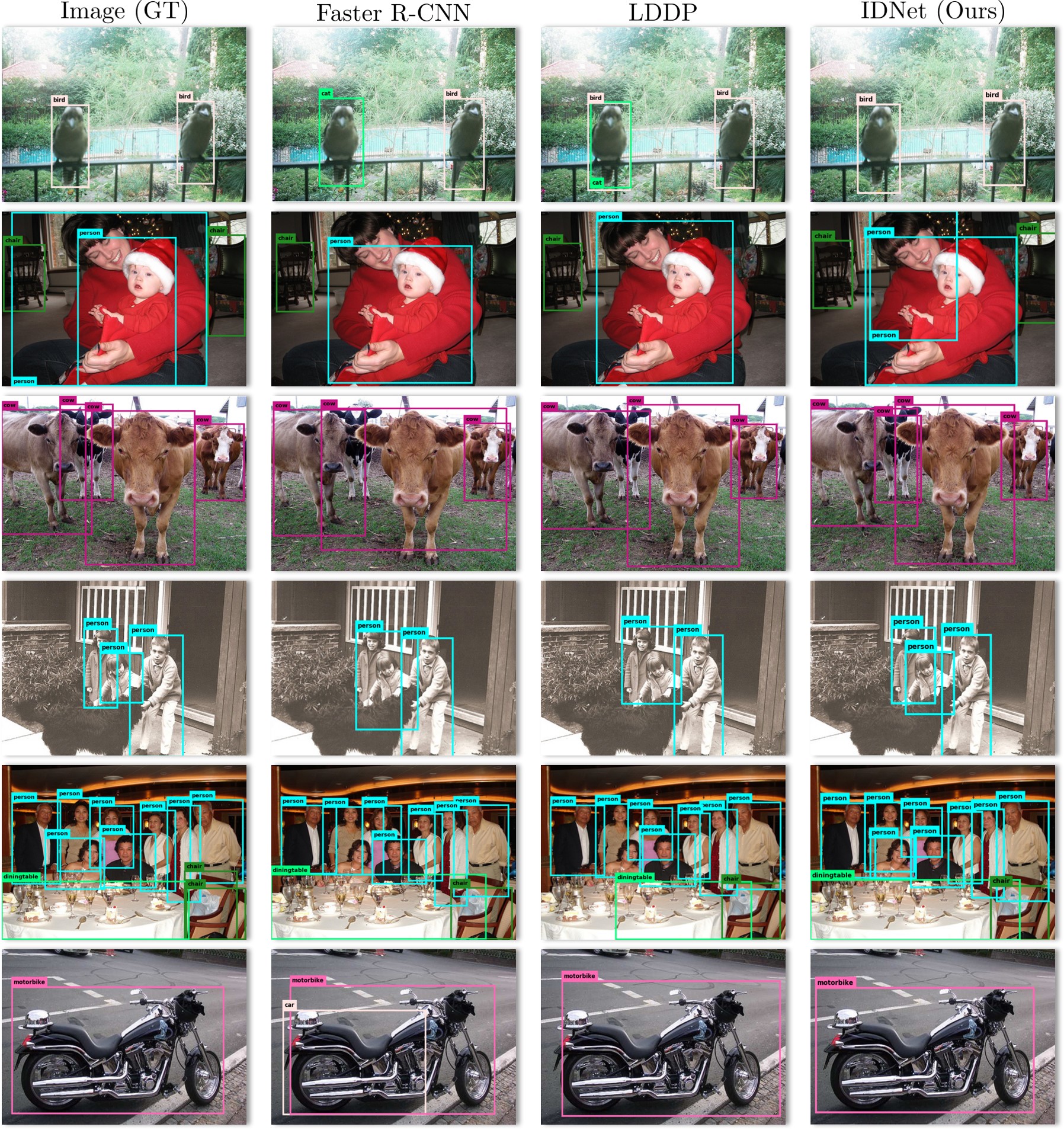}}
    \caption{\textbf{Visualization results on PASCAL VOC 2007 \texttt{test} set.} The leftmost column shows the ground truth boxes, and the other columns show the results of Faster R-CNN, LDDP, and IDNet from left to right. For each method, final boxes with scores over 0.6 are visualized on each image. All methods are trained on VOC 2007 \texttt{trainval} set using VGG-16 as a backbone.
    } \label{fig:visualize_pascal}
\end{center} \expandup \expandup \expandup \expandup
\end{figure*}

\begin{figure*}[!ht]
\begin{center}
    {\includegraphics[width=\linewidth]{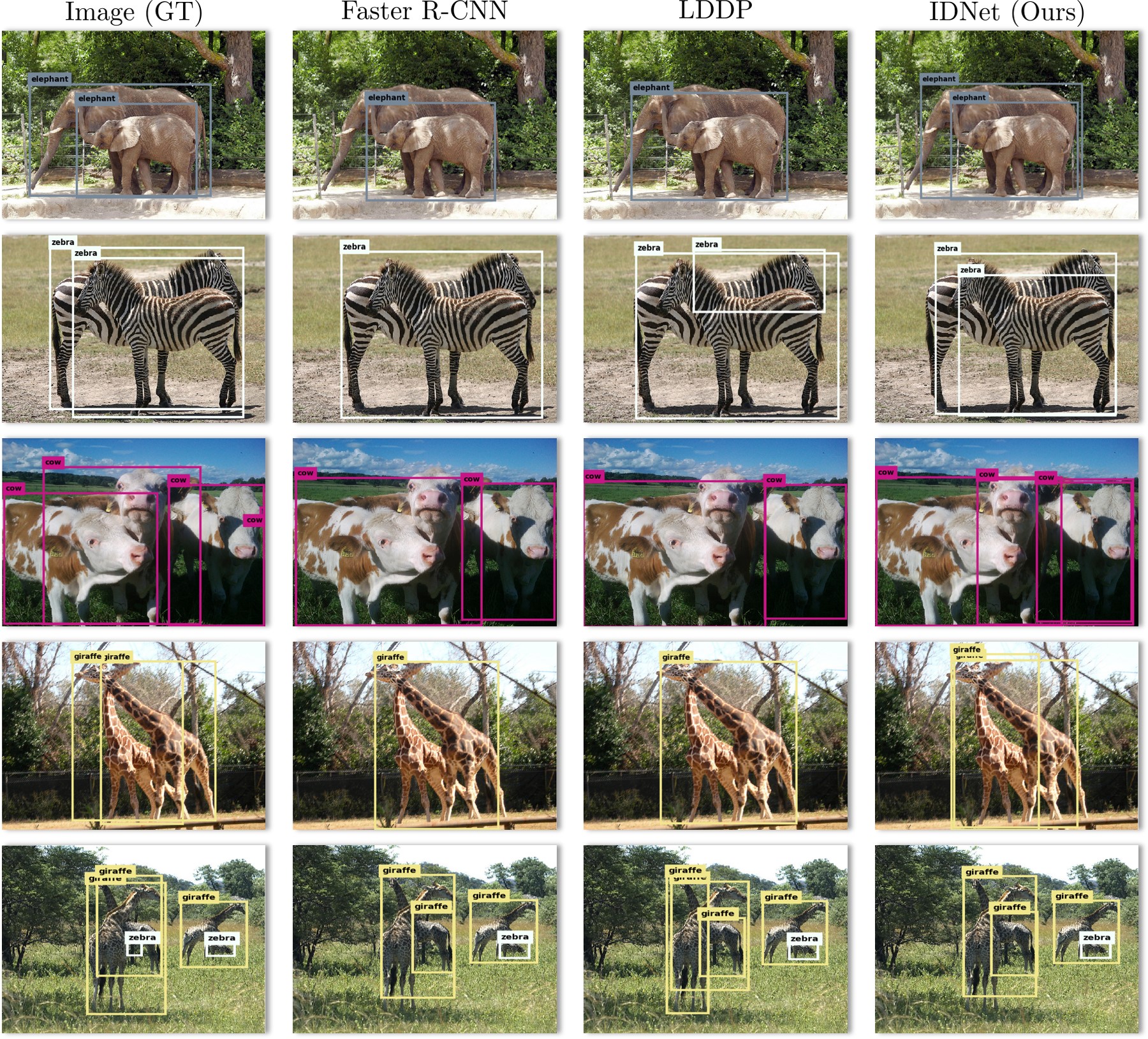}}
    \caption{\textbf{Visualization results on COCO \texttt{val} set.} The leftmost column shows the ground truth boxes, and the other columns show the results of Faster R-CNN, LDDP, and IDNet from left to right. For each method, final boxes with scores over 0.6 are visualized on each image. All methods are trained on COCO \texttt{train} set using VGG-16 as a backbone.
    } \label{fig:visualize_coco2}
\end{center} \squeezeup
\end{figure*}

\paragraph{Successful cases.}

The successful images of IDNet are visualized in Figure~\ref{fig:visualize_pascal} for VOC 2007, and Figure~\ref{fig:visualize_coco2} for MS COCO. In Figure~\ref{fig:visualize_pascal}, the first and last row images show that incorrect class bounding boxes are suppressed while selecting a correct class, which means that the IDNet suppressed bounding boxes with incorrect categories. The results on the other rows show the objects in proximity are detected while other methods fail. The results show that overlapped objects are successfully detected in IDNet. In Figure~\ref{fig:visualize_coco2}, all results show that IDNet can detect overlapped objects.

The results show the proposed IDNet can detect overlapped objects compared to the other algorithms while suppressing bounding boxes with incorrect categories.

{\small
\bibliographystyle{ieee}
\bibliography{egpaper}
}

\end{document}